\documentclass[letterpaper, 10 pt, journal, twoside]{IEEEtran}

\usepackage{graphics} 
\usepackage{epsfig} 
\usepackage{amsmath} 
\usepackage{amssymb}  
\usepackage{amsfonts}
\usepackage{subfiles}
\graphicspath{{figs/}{../figs/}}
\usepackage{csquotes}
\usepackage{gensymb}
\usepackage{pifont}
\usepackage{subcaption}
\usepackage{textcomp}
\usepackage[table]{xcolor}
\usepackage[english]{babel}
\usepackage[export]{adjustbox}
\usepackage{pdfpages}
\usepackage[font={footnotesize}]{caption}
\usepackage{blindtext}
\usepackage{soul}
\usepackage{siunitx}
\usepackage{hyperref}
\usepackage{multirow}

\def\fps@figure{htp}
\def\fps@table{htp}

\newcommand{\bi}{\begin{itemize}}
\newcommand{\ei}{\end{itemize}}

\newcommand{\bfig}{\begin{figure}}
\newcommand{\efig}{\end{figure}}

\newcommand{\benum}{\begin{enumerate}}
\newcommand{\eenum}{\end{enumerate}}

\newcommand{\be}{\begin{equation}}
\newcommand{\ee}{\end{equation}}

\newcommand{\ba}{\begin{eqnarray}}
\newcommand{\ea}{\end{eqnarray}}

\renewcommand{\deg}{${}^\circ$}

\usepackage{color}

\newcounter{magicrownumbers}

\newcommand{\figspacecaption}{\vspace{-10pt}}
\newcommand{\figspacebefore}{\vspace{0pt}}
\newcommand{\figspaceafter}{\vspace{-5pt}}
\newcommand{\tabspacecaption}{\vspace{-5pt}}
\newcommand{\tabspacebefore}{\vspace{0pt}}
\newcommand{\tabspaceafter}{\vspace{-12pt}}
\newcommand{\marginspace}{\vspace{3pt}}

\definecolor{CommentRed}{rgb}{1.0,0,0}
\definecolor{CommentBlue}{rgb}{0,0,0.7}

\title{End-to-End Velocity Estimation For Autonomous Racing}
\author{Sirish Srinivasan$^{1}$, Inkyu Sa$^{2}$, Alex Zyner$^{1}$, Victor Reijgwart$^{1}$, Miguel I. Valls$^{3}$ and Roland Siegwart$^{1}$
\thanks{Manuscript received: February 24, 2020; Revised June 18, 2020; Accepted July 19, 2020.}
\thanks{This paper was recommended for publication by Editor Jonathan Roberts upon evaluation of the Associate Editor and Reviewers’ comments.
}
\thanks{$^{1}$Authors are with the Autonomous Systems Lab, ETH Z\"urich, Z\"urich}%
\thanks{$^{2}$Author is with Robotics and Autonomous Systems Group, CSIRO, Pullenvale, Australia}%
\thanks{$^{3}$Author is affiliated with Sevensense Robotics A.G, Z\"urich}%
\thanks{Digital Object Identifier (DOI): see top of this page.}
}

\begin{document}

\markboth{IEEE Robotics and Automation Letters. Preprint Version. Accepted July, 2020}
{Srinivasan \MakeLowercase{\textit{et al.}}: End-to-End Velocity Estimation For Autonomous Racing} 

\maketitle

\begin{abstract}
Velocity estimation plays a central role in driverless vehicles, but standard and affordable methods struggle to cope with extreme scenarios like aggressive maneuvers due to the presence of high sideslip. To solve this, autonomous race cars are usually equipped with expensive external velocity sensors.
In this paper, we present an end-to-end recurrent neural network that takes available raw sensors as input (IMU, wheel odometry, and motor currents) and outputs velocity estimates. The results are compared to two state-of-the-art Kalman filters, which respectively include and exclude expensive velocity sensors. All methods have been extensively tested on a formula student driverless race car with very high sideslip (10\degree at the rear axle) and slip ratio ($\approx 20$\%), operating close to the limits of handling. The proposed network is able to estimate lateral velocity up to 15x better than the Kalman filter with the equivalent sensor input and matches (0.06 m/s RMSE) the Kalman filter with the expensive velocity sensor setup.
\vspace{-4pt}
\end{abstract}

\IEEEpeerreviewmaketitle

\vspace{-8pt}
\section{Introduction}\label{sec:introduction}
\vspace{-2pt}
\IEEEPARstart{S}{elf-driving} cars have become popular in recent years because they promise to transform cities, provide universal access to mobility, and increase transport efficiency \cite{Litman2020-gi}. However, to reach full (level 4) autonomy as defined by the Society of Automation Engineers (SAE)~\cite{cit:sae}, no driver attention must be required even in extreme scenarios or unusual conditions like adverse weather.  Most research focuses on maneuvers that can be modeled kinematically since these cover most common cases. Dynamic maneuvers that imply high sideslip can arise in emergency avoidance maneuvers, in which the autonomous pipeline must also remain functional. Therefore, autonomous racing presents an interesting scenario to test all algorithms at the limit of handling.

\begin{figure}
\figspacebefore
\begin{center}
\includegraphics[width=0.68 \columnwidth]{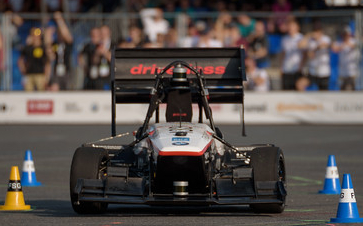}
\vspace{-2pt}
\includegraphics[width=0.9 \columnwidth]{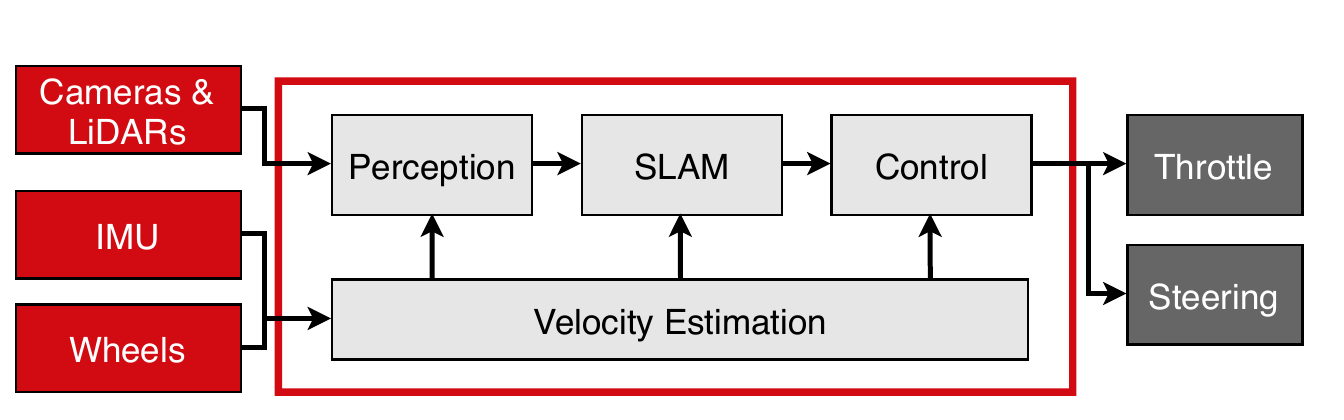}
\end{center}
    \figspacecaption
	\caption{Top: \emph{pilatus} driverless, the formula student race car used for testing. \copyright FSG - Zenker. Bottom: Simplified software architecture of \emph{pilatus} driverless race car showing the central role of velocity estimation}
	\label{fig:pilatus_driving_and_architecture}
\figspaceafter
\vspace{-7pt}
\end{figure}

Velocity estimation has a central and pivotal role in the entire autonomous system. The velocity estimates are for example used by the perception pipeline to perform motion undistortion on the LiDAR pointclouds, by the SLAM pipeline as the motion model, and by the vehicle motion control system for the critical task of providing state feedback. Velocity estimation must therefore provide high-rate, high-quality data. The overall architecture of similar autonomous race cars has been described in detail by Kabzan \textit{et al.} \cite{cit:amz_full} and a simplified diagram is shown in Figure~\ref{fig:pilatus_driving_and_architecture}. 

\vspace{-5pt}
The standard velocity estimation systems can be classified into two categories, both based on Kalman filters. The first type is often used in Electronic Stability Program (ESP) modules of road cars, like \cite{cit:ve_review, cit:vy_review}, and does not have external velocity sensors. Such systems are designed to work in standard driving conditions but fail in extreme ones. The second type of velocity estimation systems relies on expensive sensors and is typically used on race cars. These systems prevail in extreme conditions, but cannot cost-effectively be deployed on commercial road cars~\cite{cit:fluela_icra, cit:gotthard_icra, cit:tum_ve}. 

In this paper, we propose a learning-based method to perform velocity estimation in extreme scenarios like racing (second category) without external velocity sensors. This technique achieves equivalent performance to Kalman Filters that include external velocity sensors. The method is demonstrated on a full scale autonomous racecar (Figure \ref{fig:pilatus_driving_and_architecture}) which is able to accelerate from $0-100 km/h$ in $2.1s$ and reaches lateral accelerations of $1.7g$.

The contributions of this paper are as follows:
\vspace{-2pt}
\begin{itemize}
\item Presenting a novel recurrent neural network application to estimate the velocity of a car using only raw, inexpensive sensor measurements (e.g., IMUs, wheel odometry and motor currents).
\item Intensive real-world performance evaluation of the proposed approach and comparison with Kalman Filter approaches.
\item Showing how the proposed end-to-end estimation approach outperforms state-of-the-art Kalman Filters with equivalent sensor setups by a large margin, and matches the approaches with external velocity sensors.
 
\end{itemize}

Section \ref{sec:related work} presents related work. The proposed network, and Kalman filter are presented in Section (\ref{sec:methodology}), followed by real-world experimental results in Section \ref{sec:experiments}. The limitations and conclusions are presented in Sections \ref{sec:discussion} and \ref{sec:conclusion}.

\section{Related work}\label{sec:related work}In this section, we start by reviewing conventional state estimation approaches (e.g., Stochastic Filtering) and then move on to deep learning-based approaches from which we drew inspiration.

Xue and Schwartz \cite{cit:filters} compare the state-of-the-art approaches for state estimation in wheeled mobile robot applications. The main differences between these approaches and race cars are the complex dynamics that occur when driving at the limits of handling ($18 m/s$ and $1.5g$ lateral acceleration), which are very difficult to model completely. Wheel speeds are a biased measurement of the velocity, due to the high lateral and longitudinal slips (up to 20\%) that race cars require to reach optimal accelerations as described by Pacejka \cite{cit:Pacejka2012}.

An Extended Kalman Filter (EKF) is used by Valls \textit{et al.} \cite{cit:fluela_icra} to fuse data from multiple sensors and perform velocity estimation for an autonomous race car. Outlier detection and observability analysis have been studied as a part of this work to find the minimal sensor setup. Gosala \textit{et al.} \cite{cit:gotthard_icra} extend this work by using a complex slip ratio-based model to perform reliable velocity estimation even during sensor failure. The only drawback of this approach is the dependence on dedicated velocity sensors to provide a reliable estimate. The model mismatch at high speeds and during aggressive maneuvers causes the estimate not to be as accurate. Wischnewski \textit{et al.} \cite{cit:tum_ve} describe an EKF to fuse data from a velocity sensor and lidar data for velocity estimation in a high-speed roborace car. Even they discuss a reduction in accuracy of the estimate during velocity sensor failure. 


A deep Rectified Linear Unit (ReLU) based neural network has been used by Punjani and Abbeel \cite{cit:heli_dynamics} to model the complex dynamics of a helicopter during aggressive maneuvers. They show that this model outperforms state-of-the-art methods by more than 50\%. Spielberg \textit{et al.} \cite{cit:stan_nndyn} use a deep neural network with delayed inputs over multiple time steps to model the dynamics of a high-performance car and have shown it works over a variety of operating conditions. These approaches motivate the usage of neural networks to model complex dynamics that would not be possible with traditional approaches. Jonschkowski \textit{et al.} \cite{cit:diff_part_filt} show the benefits of combining end-to-end learning with algorithmic priors by applying their differentiable particle filters to global localization and visual odometry tasks.

Karpathy \cite{cit:karpathy_rnn} demonstrates the effectiveness of Recurrent Neural Networks (RNNs) in time series prediction. It has been shown to learn the time dependency of the underlying dynamics without explicitly defining the number of input time steps. Only one time step is used as input every time step and the hidden state is propagated over time. This reduces the need to optimise for the number of input time steps to the network during deployment.

One drawback of RNNs is the vanishing gradient problem during back propagation which has been discussed in detail by Pascanu \textit{et al.} \cite{cit:rnn_gradients}. To counter this, Hochreiter \textit{et al.} \cite{cite:lstm_original} use Long-Short Term Memory (LSTM) cells to build the RNN. This allows the network to propagate gradients over time and learn long-term dependencies. Jozefowicz \textit{et al.} \cite{cit:gru_vs_lstm} compare the performance of Gated Recurrent Units (GRU) and LSTM cells. They show that LSTMs are harder to train and are dependent on the chosen network architecture. Also, an LSTM network has more weights to optimise when compared to a GRU, requiring more data to train. 

Drews \textit{et al.} \cite{cit:autorally_lstm} use a mono camera processed using CNNs combined with LSTMs to obtain a representation of the track. This is then combined with IMU, wheelspeeds and RTK GNSS using a particle filter to estimate the states. 
Shashua \textit{et al.} \cite{cit::deep_kf} present a Deep Robust Kalman Filter for Robust Markov Decision Processes, which accounts for uncertainty in the weights of the value function and the transition probabilities.

Overall, modelling the dynamics of a vehicle involves the identification of many parameters, most of which have a very minor effect on the actual dynamics. Also, extracting meaningful information from noisy data requires extensive filtering techniques. We combine both these steps and use a machine learning-based approach to perform end-to-end velocity estimation from the sensor measurements to state estimates. To the best of the authors' knowledge, this approach hasn't been done before in the literature.

This is then tested on real sensor data obtained on a formula student driverless car, and the performance is compared to the Kalman Filter reference.

\section{Methodology}\label{sec:methodology}In this section, we present two approaches to estimate the car's velocity: a classic filtering approach (Mixed Kalman Filter), and a learning-based approach (End-to-end Recurrent Neural Network).
\subsection{Sensors}\label{sec:sensors}
\begin{figure}
\figspacebefore
\marginspace
\begin{center}
\includegraphics[width=\columnwidth]{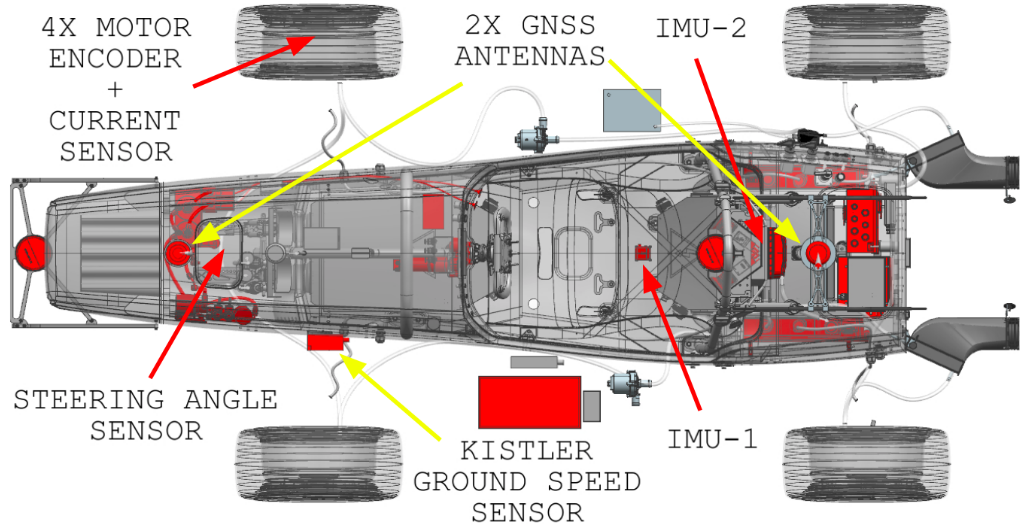}
\end{center}
    \figspacecaption
	\caption{Sensor setup: two IMUs, 4x motor encoders + current sensors, steering angle sensor and two dedicated velocity sensors that are used for validation and target generation}
	\label{fig:sensor_setup}
\figspaceafter
\vspace{-5pt}
\end{figure}

The sensor setup, shown in Figure \ref{fig:sensor_setup}, includes two IMUs measuring accelerations (x,y) and rotational rates (z), four motor encoders measuring the wheel speeds, steering angle sensor and motor torques of each wheel (derived from the motor current sensors). There are two velocity sensors (an optical flow-based velocity sensor and a GNSS velocity sensor) which are used only for validation and target generation. 
\vspace{-5pt}

\subsection{Mixed kalman filter}\label{sec:Kalman_filter}
As a baseline and the first approach to estimate velocity, we develop a Kalman Filter to fuse all the sensor measurements. The presented propagation model (Eq.~\ref{eq:propagation_model}) is mildly non-linear, and all measurement models are linear except for the combined velocity measurement (Eq.~\ref{eq:sensor_model}), which is highly non-linear. For this reason, and to render the filter as computationally light as possible, a mix of the Linear Kalman filter (LKF), Extended Kalman filter (EKF) and Unscented Kalman Filter (UKF) is used. The Mixed Kalman Filter (MKF) propagates the state and covariance with an EKF step, the rotation and acceleration measurement with LKF updates, and only the combined velocity measurement with a UKF update. This proved to increase the accuracy of the filter with reduced computational load, beating state-of-the-art filters such as~\cite{cit:gotthard_icra}.

\subsubsection{States and propagation model}\label{sec:state_and_propagation}
The system's state, $\mathbf{x}$, is defined as:
\begin{align}\label{eq:state}
\begin{split}
\mathbf{x} &= [\mathbf{v}^T, \,\dot{\psi}, \, \mathbf{a}^T]^{T}, \\
\mathbf{v} &= [v_x, \, v_y]^T, \, \mathbf{a} = [a_x, \, a_y]^{T}, \\
\end{split}
\end{align}
where $\mathbf{v}$ is the velocity, $\dot{\psi}$ the rotational rate along the z-axis (i.e., yaw) and $\mathbf{a}$ the acceleration. The system evolution can described by:
\begin{align}\label{eq:propagation_model}
\begin{split}
\dot{\mathbf{v}} &= \mathbf{a} + \left[ v_y \dot{\psi}, \, -v_x \dot{\psi} \right]^T+ \mathbf{n_v},\\
\ddot{\psi} &= n_r, \, \dot{\mathbf{a}} = \mathbf{n_a},
\end{split}
\end{align}
where  $n_{\{\cdot\}}$ are white i.i.d noises.

\subsubsection{Measurement model}
The measurement models of rotation rates, and accelerations are straight forward since they belong to the state. For the other sensors, they are combined into one single velocity measurement as follows: 
\begin{align}\label{eq:sensor_model}
\mathbf{z_{v}} &= h_{v}(\mathbf{x}) = (\mathbf{v_{\text{min}}} + [-\dot{\psi}\, p_{y}, \, \dot{\psi}\, p_{x}]^T) + \mathbf{n_{z_{vx}}}\\
\mathbf{v_{i}} &= [cos(\delta_i), \, sin(\delta_i)]^{T} \cdot \omega_{i} \cdot R_i / (SR(T_i)+1) + n_{z_{v}} 
\end{align}
$\mathbf{z_{v}}$ is the combined velocity measurement, $\mathbf{v_i}$ is the velocity of the \textit{ith} wheel in car frame, $\omega_{i}$ is the rotational velocity of the \textit{ith} wheel, $\delta$ is the angle of each wheel w.r.t car frame, which is the steering angle for the front and $0$ for the rear wheels. $SR(\cdot)$ is a function that maps the torque applied on a wheel ($T_i$) to its slip ratio under low slip conditions (Pacejka magic tire model \cite{cit:Pacejka2012}). $R_i$ is the radius of the \textit{ith} wheel. $p_x$ and $p_y$ is the wheel position in car frame.\\
Slip ratio calculation is fairly accurate at low slips, but uncertain at high slips, this is why in practice, the velocity update only takes the single wheel with smallest absolute SR. The linear velocity of this wheel is denoted $\mathbf{v_{\text{min}}}$ 

\subsection{End-to-End Recurrent Neural Network}\label{sec:end_to_end}
The core of this work is a recurrent end-to-end learning approach. The temporal nature of the data makes recurrent neural networks particularly suitable. Two networks are presented and compared, each being the best performing hyperparameter combination with 1 and 2 GRU layers, denoted RNN-1 and RNN-2, respectively.

\subsubsection{Target calculation}\label{sec:target_calculation}
To generate the target for the RNN, an MKF was used (See Section ~\ref{sec:Kalman_filter}) in combination with the sensors described in Section~\ref{sec:sensors} including the external velocity sensors. This filter was extensively tested and proved to be successful in multiple Formula Student competitions around Europe in 2019. The target of the end-to-end approach should thus obtain similar results but do so without relying on expensive velocity sensors. The output from the MKF is post-processed using a non-causal Gaussian moving average filter to obtain a smoothed non-delayed target referred to as the reference.

\subsubsection{Network architecture}
\begin{figure}
\figspacebefore
\marginspace
\begin{center}
\includegraphics[width=0.95\columnwidth]{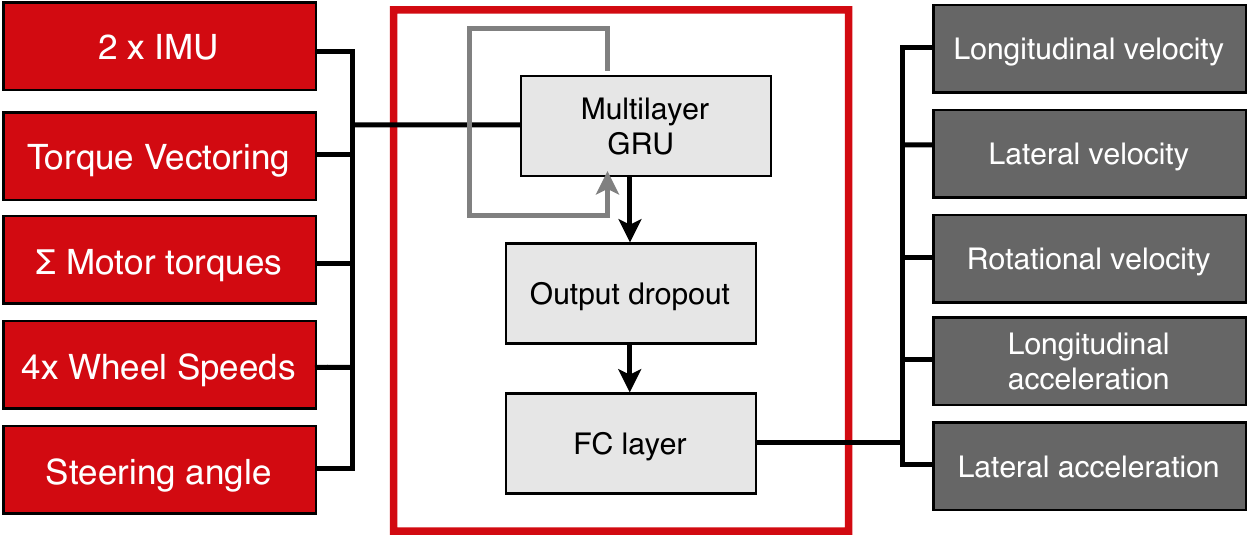}
\end{center}
    \figspacecaption
	\caption{Basic architecture of the proposed recurrent neural network consisting of multiple hidden GRU layers that propagate over time, followed by an output dropout and a fully connected dense layer to extract the outputs.} \label{fig:rnn_architecture}
\figspaceafter
\vspace{-8pt}
\end{figure}

The inputs to the network are the sensors described in Sec.~\ref{sec:sensors}, and the overall dimension of the input vector is 13 (3 per IMU, 2 torques, 4 wheel speeds, and 1 steering angle). The output is the state estimate, described in Sec.~\ref{sec:state_and_propagation}. Different architectures were explored, and two were selected, named RNN-1 and RNN-2 and shown in Figure \ref{fig:rnn_architecture}. RNN-1 has a single hidden layer of 64 GRU neurons while RNN-2 has two hidden layers of 32 neurons each in the multilayer GRU. The rest of the architecture remains the same for both the networks, and includes a dropout layer and a fully connected dense layer, to transform the inner network state to the outputs.

\subsubsection{Activation function}
Pascanu \textit{et al.} \cite{cit:rnn_gradients} discuss the problem of vanishing gradient for RNNs, especially when the activation function used saturates for high inputs - like sigmoid and tanh. To overcome this, Xu \textit{et al.} \cite{cit:leaky_relu} suggest using a leaky-ReLU which does not saturate for high inputs and has a small negative output for negative inputs. This ensures that cells do not "die" and that they participate in the training even after being wrongly modified by the optimiser.

\subsubsection{Optimiser}
The main trade-off for an optimiser lies between training time and stability. Two common optimisers that implement adaptive learning rate are Adadelta \cite{cit:adadelta} and Adam \cite{cit:adam_optimiser}. Training with Adadelta consistently converged irrespective of network architecture, probably due to its lack of tuning parameters. On the other hand, Adam's learning rate needs to be tuned to find this trade-off. However, once a good learning rate has been identified ($5\times10^{-4}$), Adam slightly outperforms Adadelta and is therefore chosen.

\subsubsection{Output dropout}
Dropout is added to only the output layers, and not the recurrent layers as per Zaremba \textit{et al.} \cite{cit:rnn_dropout}. Output dropout is applied only during training and not during testing and validation. A dropout fraction of $7.5$\% was found to have the similar train, test and validation losses.

\section{Experiments and Analysis} \label{sec:experiments}In this section, we refer to the MKF with and without velocity sensors as reference and baseline, respectively. We refer to RNN-1 and RNN-2 as described in Sec. ~\ref{sec:end_to_end}.

\subsection{Data collection}
Datasets were created from real data obtained from noisy sensors over both testing and competition runs on Formula Student style tracks. The raw sensor measurements have different sampling frequencies of 200Hz (IMU1 and steering angle), 125 Hz (IMU2), and 100Hz (Wheel speeds and torques). These raw measurements are hardware time-synced by sampling at a constant rate of 200 Hz (zero-order hold) and used as input to the network. The target is generated as described in Sec.~\ref{sec:target_calculation}. The datasets include data from different road surfaces -- flat (35\%), gravel (29\%), bumpy (27\%) and wet (9\%) -- along with variation in temperatures (20-60\deg{}C) and grip conditions. This has been split into training, testing, and validation sets, as shown in Table \ref{tab:dataset_table}. Each of these splits has data from the various road conditions described above.

\begin{table}[]
\tabspacebefore
\vspace{8pt}
\begin{center}
 \begin{tabular}{|l|c|c|}
 \hline
 Dataset Type & \# Datasets & Duration \\ \hline \hline
 Training & 11 & 22 min 28.440 s\\
 Testing & 3 & 5 min 8.485 s \\
 Validation & 4 & 4 min 40.990 s \\
 Total & 17 & 32 min 17.915 s \\
 \hline
 \end{tabular}
 \tabspacecaption
  \caption{Dataset split between training, testing and validation for the recurrent neural network}
 \label{tab:dataset_table}
\end{center}
\tabspaceafter
\end{table}

\subsection{Implementation details}
\begin{figure}
\figspacebefore
\begin{center}
\includegraphics[width=0.95\columnwidth]{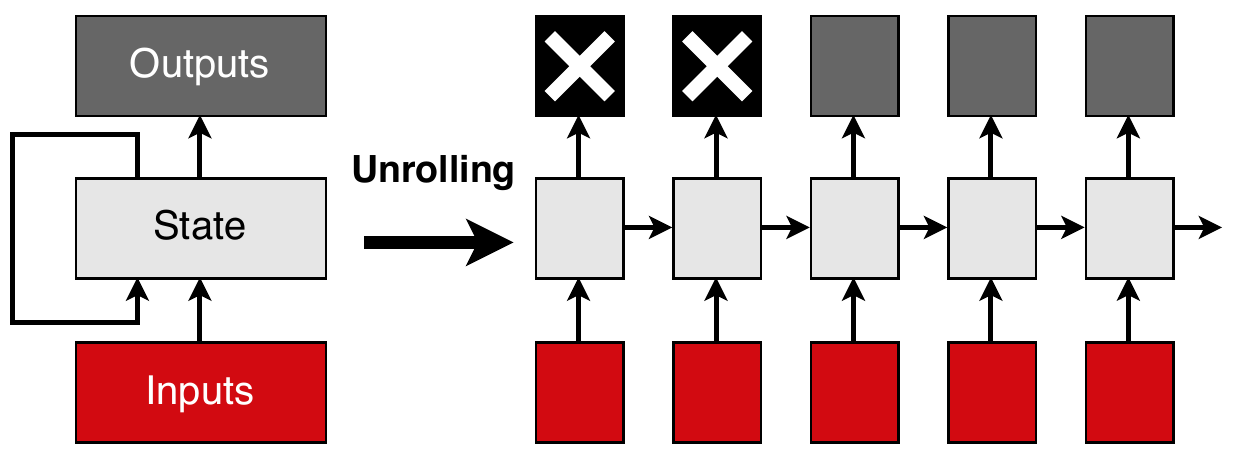}
\end{center}
    \figspacecaption
 	\caption{Pictorial representation of a training sample with the inputs and outputs in each time step. The hidden state propagates over time providing the possibility to learn dependencies in the time-series data. The output for the initial time steps (shown in black, crossed out) are ignored while calculating losses.}
	\label{fig:rnn_unrolling}
\figspaceafter
\vspace{-8pt}
\end{figure}

The internal hidden states of the RNN learn dependencies between the time series data, shown graphically in Figure \ref{fig:rnn_unrolling}. The RNN predictions for the first few time steps are ignored and the prediction loss computed as the root mean square error (RMSE) after this initial setup period. This period is chosen as $200$ samples or $1s$, which provides a nice trade off between output quality and initial wait time.\\
The training is performed in parallel using a batch size of 500 steps, and including standard methods like input-output normalisation, gradient clipping and early stopping. The network was implemented in Tensorflow using the dynamic\_rnn package for unrolling over time. Multiple GRU layers are stacked together using the Multi RNN Cell package.\\
The prediction time of the network on a Nvidia GeForce RTX 1080 Ti GPU for 500 timesteps is 0.20 ms.

\subsection{Hyperparameter study}
\begin{figure}
\figspacebefore
\marginspace
\begin{center}
\includegraphics[width=0.95\columnwidth]{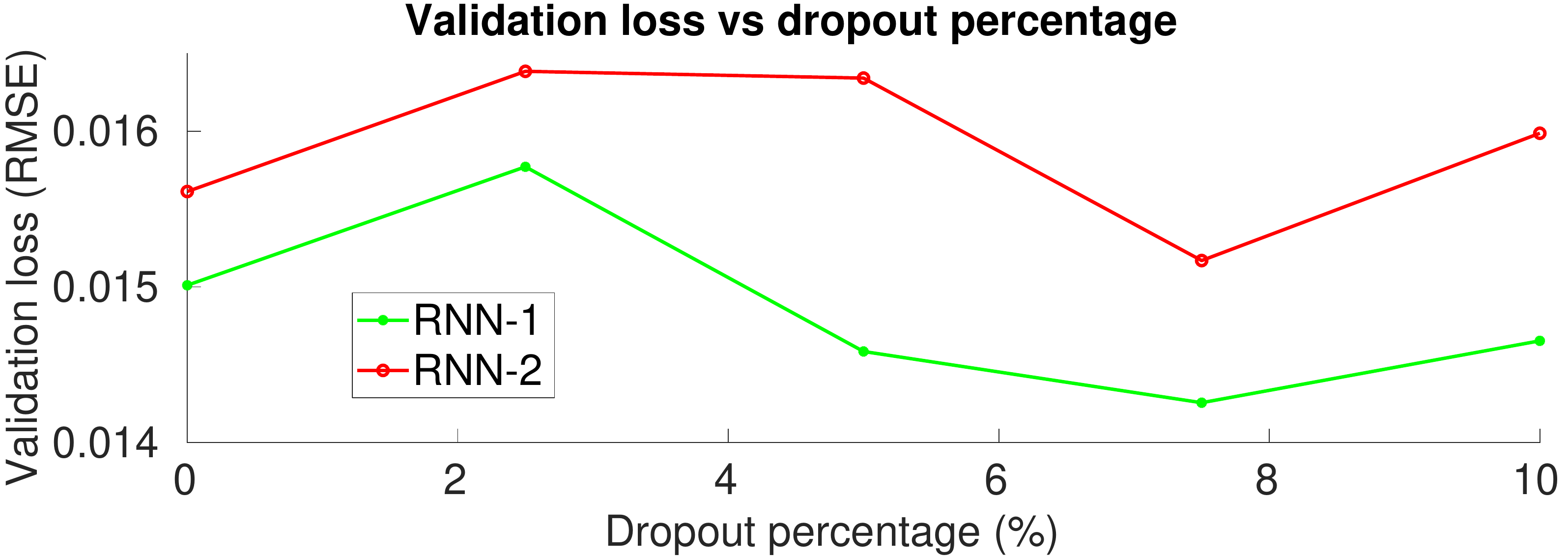}
\end{center}
    \figspacecaption
 	\caption{Validation loss as a function of output dropout percentage. The dropout percentage with minimum validation loss is 7.5\% for both networks}
	\label{fig:valid_dropout}
\figspaceafter
\end{figure}

The different hyper-parameters that were searched are listed in Table \ref{tab:sensitivities}, and the final values are shown in bold. The performance variation for one such hyper-parameter, the output dropout percentage, is shown in Figure \ref{fig:valid_dropout}, and it can be seen that 7.5\% has the lowest validation loss.

\begin{table}[]
\tabspacebefore
\begin{center}
 \begin{tabular}{|l|c|c|}
 \hline
Hyper-parameter & Searched values \\ \hline 
Number of GRU layers & \textbf{1}, 2, 3\\
Neurons per layer & 16, \dots \textbf{64} \dots256\\
Input steps for training sample & 20\dots \textbf{300}\dots500\\
Output steps for training sample & 100\dots\textbf{200}\dots1000\\
Activation function & relu, \textbf{leaky-relu}, elu\\
Optimiser & adadelta, \textbf{adam}\\
Learning rate & 0.0001\dots\textbf{0.0005}\dots0.01\\
Gradient Clipping - global norm & No, \textbf{Yes}\\
Dropout fraction & 0\dots\textbf{0.075}\dots0.25\\
Training epochs & 1,000\dots\textbf{10,000}\\
 \hline
 \end{tabular}
 \tabspacecaption
 \caption{Hyper-parameters and the ranges that were iterated over with the final values in bold}
 \label{tab:sensitivities}
\end{center}
\tabspaceafter
\vspace{-8pt}
\end{table}

\subsection{Accuracy}
The main quantitative analysed metrics are the RMSE of the prediction (from RNN or MKF) vs the reference as well as the normalized error \%. The results can be seen in Table~\ref{tab:performance_metrics}. It can be noted that lateral velocity has the highest prediction error, and this is due to it being the hardest to estimate given the input sensors. In fact the baseline MKF has an error 15 times larger than both RNNs. 
\begin{table}[]
\tabspacebefore
\begin{center}
\begin{tabular}{|c|c|c|c|c|c|c|}
 \hline
 \multirow{2}{*}{State} & \multicolumn{2}{c|}{Baseline} & \multicolumn{2}{c|}{RNN-1} & \multicolumn{2}{c|}{RNN-2} \\
 \cline{2-7}
  &RMSE&\%Error&RMSE& \%Error&RMSE&\%Error\\ \hline
  $v_x$ & 0.779 & 5.95 & 0.141 & \textbf{0.94} & 0.159 & 1.06 \\
  $v_y$ & 0.898 & 59.78 & 0.059 & \textbf{3.90} & 0.061 & 4.09 \\
  $\dot{\psi}$ & 0.018 & \textbf{1.21} & 0.029 & 1.91 & 0.030 & 2.00 \\
  $a_y$ & 0.665 & 4.43 & 0.411 & \textbf{2.74} & 0.450 & 3.00 \\
  \hline
 \end{tabular}
 \tabspacecaption
 \caption{Quantitative Performance metrics comparing the performance of the proposed recurrent neural networks with the Kalman filter reference and baseline}
 \label{tab:performance_metrics}
\end{center}
\tabspaceafter
\end{table}

\begin{figure}
\figspacebefore
\marginspace
\begin{center}
\includegraphics[width=0.85\columnwidth]{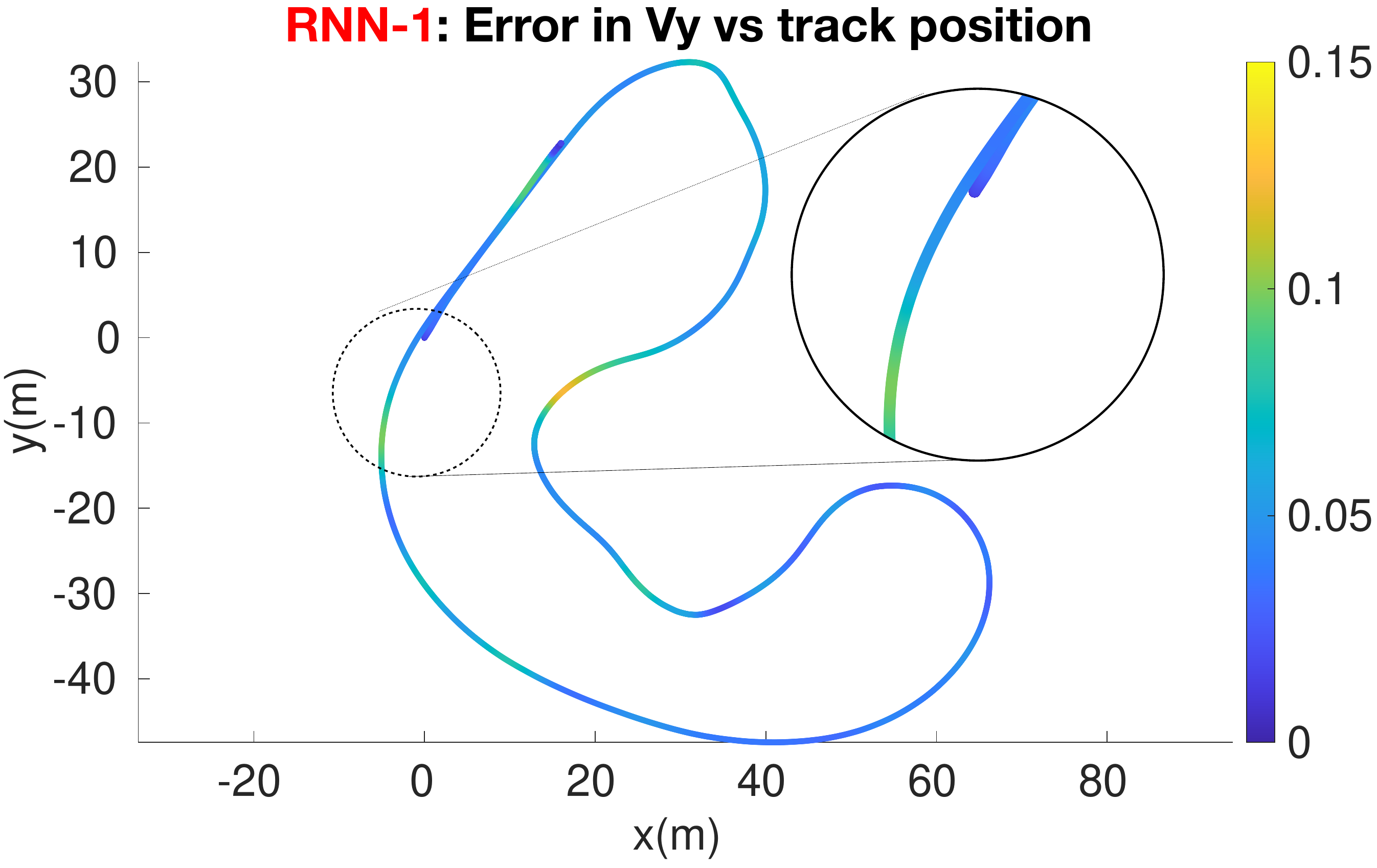}
\includegraphics[width=0.85\columnwidth]{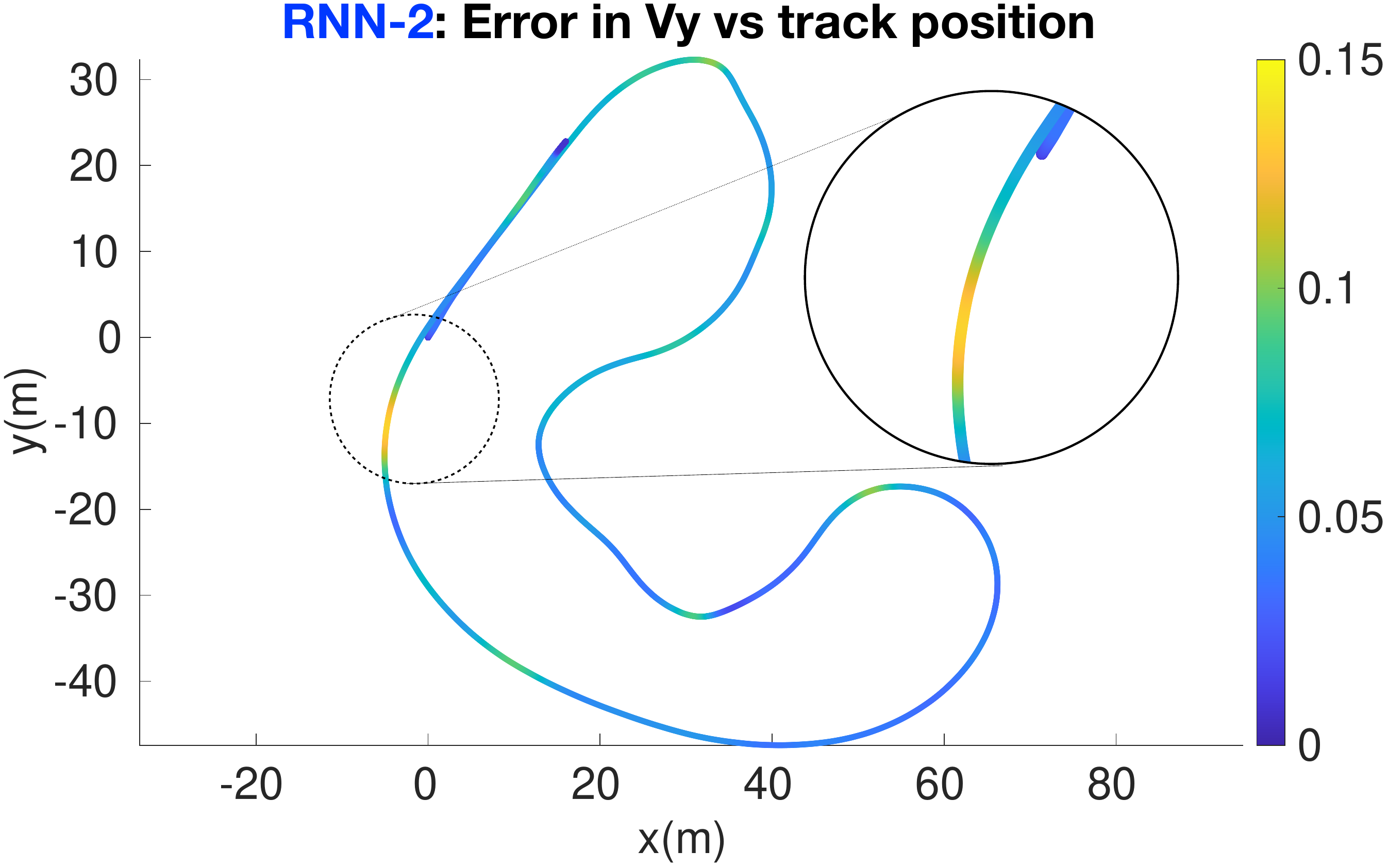}
\end{center}
    \figspacecaption
	\caption{Lateral velocity error along the track for the single layer RNN (RNN-1 in red) and double layer RNN (RNN-2 in blue). The highest error can be observed in the sharp transition from right to left for RNN-1 and in the high speed corner for RNN-2.}
	\label{fig:vy_vs_pos}
\figspaceafter
\end{figure}

\begin{figure}
\figspacebefore
\begin{center}
\includegraphics[width=0.95\columnwidth]{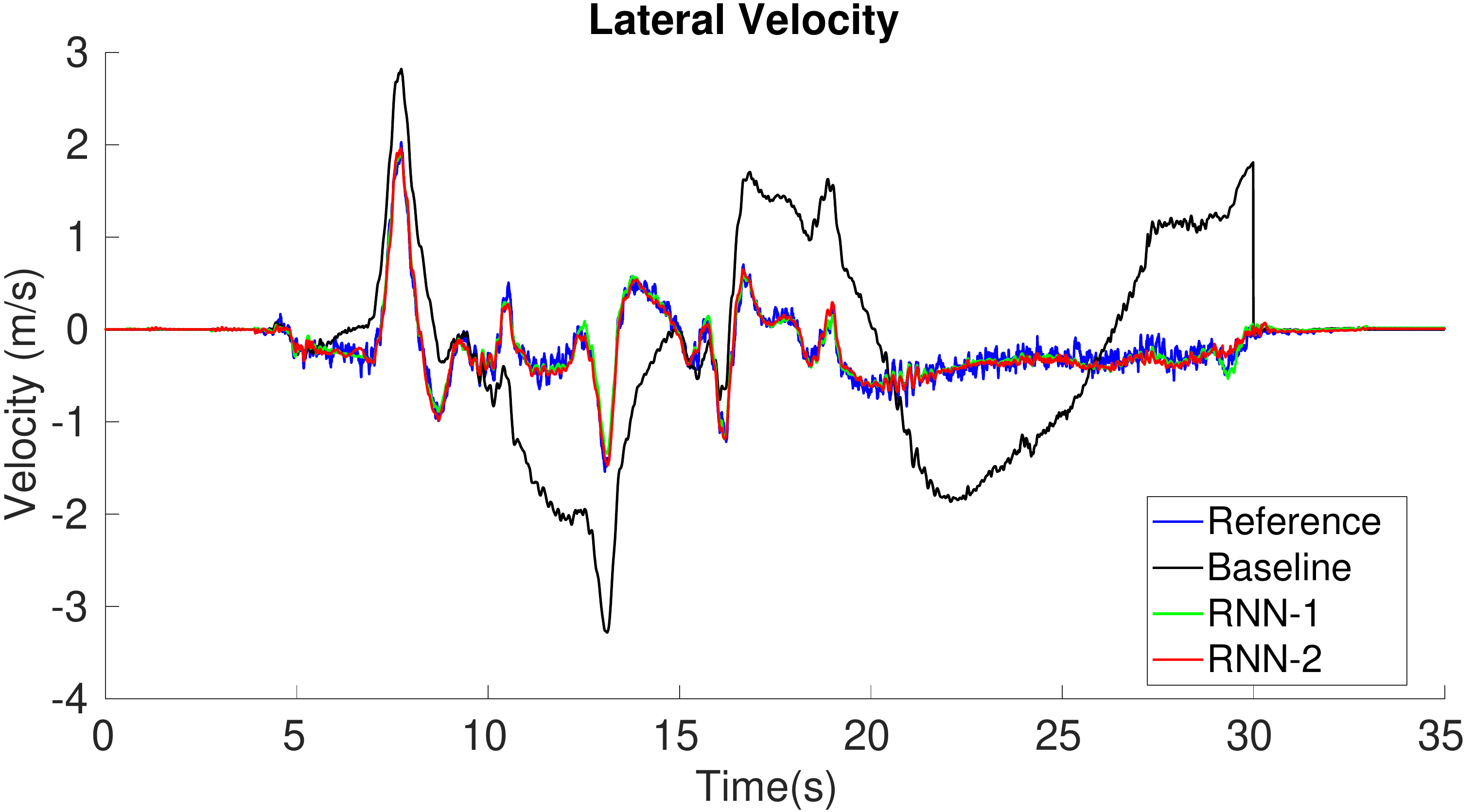}
\end{center}
    \figspacecaption
	\caption{Comparison of the lateral velocity estimates shows the superior performance of the proposed networks w.r.t the Kalman filter baseline. The predictions are comparable to the Kalman filter reference even for the hardest state to estimate.}
	\label{fig:vy_comparison}
\figspaceafter
\vspace{-12pt}
\end{figure}

In addition to the RMSE, we analyse the quality of the estimate by visual inspection. Figure \ref{fig:vy_vs_pos} shows the lateral velocity error at different positions along the track for one Formula Student Germany (FSG) run. The regions with the highest error correspond to the ones where the vehicle experiences the highest lateral slip. Figure \ref{fig:vy_comparison} shows the performance of the proposed recurrent neural networks as compared to the reference and the MKF baseline over the entire lap. The networks clearly outperform the baseline and perform similar to the reference even without using velocity sensor measurements. This highlights the dependence of the Kalman filter on a direct velocity sensor.

\subsection{Case studies}
\label{sec:case_studies}
We discuss some case studies below that highlight the performance of the proposed network in some difficult scenarios for velocity estimation. Since both the one layer and two layer architectures have similar performance, the single layer RNN is preferred due to a lower prediction time ($\approx 30$\%) and we would focus on this in the rest of the paper.

\subsubsection{Bias calibration} \label{sec:bias_calib}
\begin{figure}
\figspacebefore
\marginspace
\begin{center}
\includegraphics[width=0.95\columnwidth]{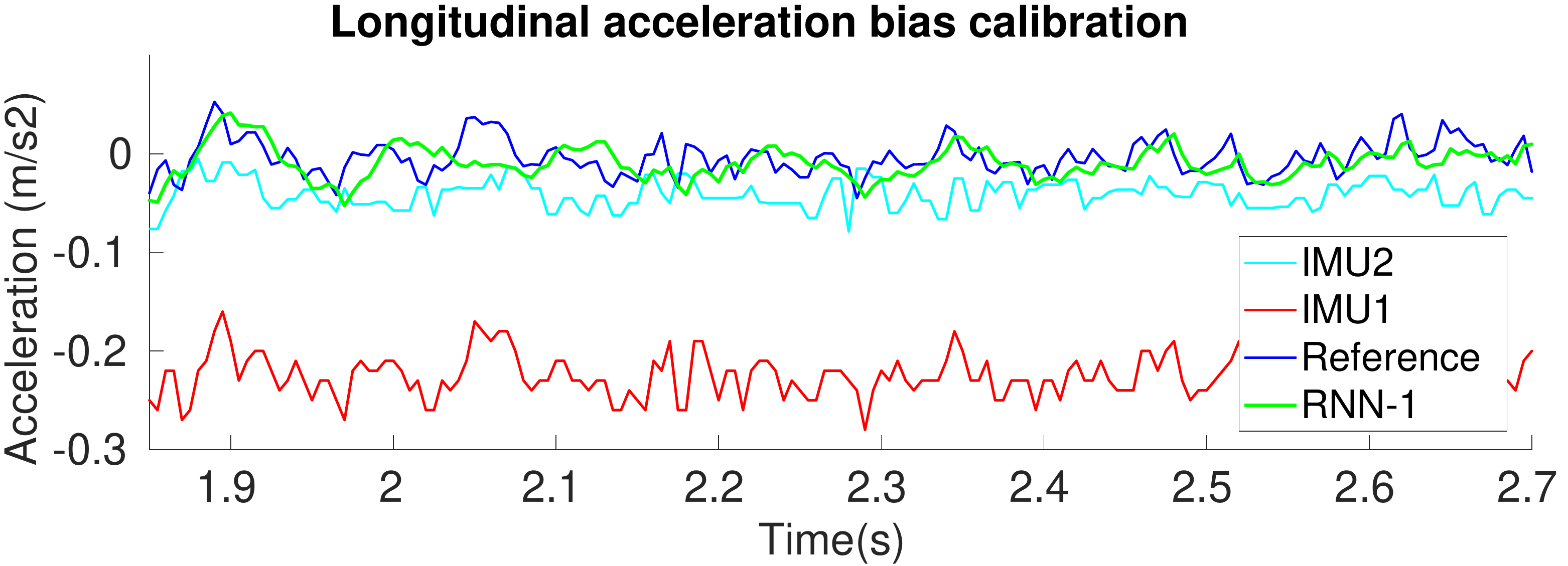}
\end{center}
    \figspacecaption
	\caption{The proposed network is able to calibrate for the bias in raw sensor measurements and provide an unbiased estimate of longitudinal acceleration}
	\label{fig:bias_calibration}
\figspaceafter
\end{figure}

For the MKF, the biases of the various sensors are calibrated explicitly by assuming that the car stands still during startup. The proposed network instead had to learn to estimate these biases, and the results can be seen in Figure \ref{fig:bias_calibration}. Two sensor measurements, one from each IMU, with different biases, are shown while the vehicle is stationary. The accelerometer biases are observable since the velocity is part of the network's input and corresponds to the integrated acceleration. As desired, the output from the network is an unbiased estimate of the acceleration centered at zero. On the other hand, the gyro biases are not observable. The prediction error of the RNN for rotational velocity is therefore larger (see Table \ref{tab:performance_metrics}), and in practice incorporating the same explicit calibration procedure as used by the MKF would be recommended.

\subsubsection{Vehicle launch}
\begin{figure}
\figspacebefore
\begin{center}
\includegraphics[width=0.95\columnwidth]{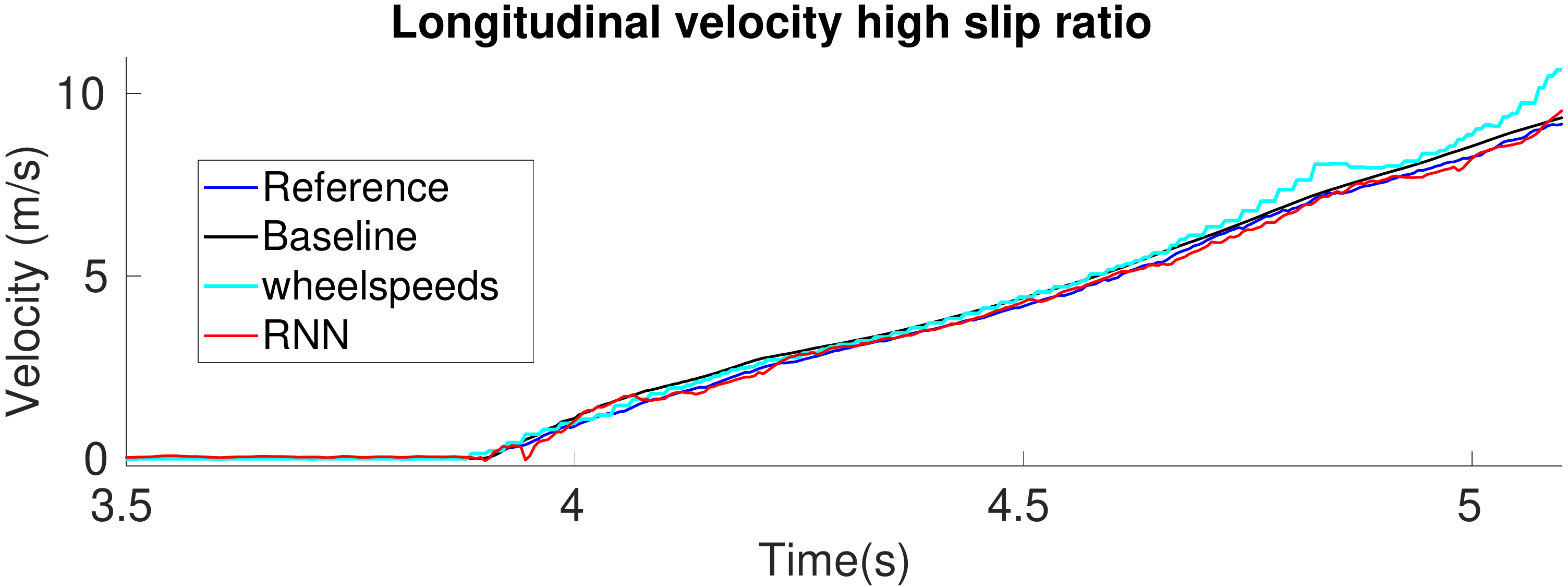}
\end{center}
    \figspacecaption
	\caption{The performance of the proposed network for the longitudinal velocity estimate is comparable to the Kalman filter reference even during vehicle launch - a very high slip ratio condition ($\approx 20$\%)}
	\label{fig:vx_launch}
\figspaceafter
\vspace{-8pt}
\end{figure}

During launch, the car accelerates from stand still using traction control to maximise its acceleration. This requires an accurate estimate of the vehicle's speed which is hard without velocity sensors since the wheels are slipping. Figure \ref{fig:vx_launch} shows the performance of the proposed network in this case. Initially the velocity is slightly over-estimated due to wheel slip, but then the network is able to identify this after a few steps and corrects it resulting in a performance similar to the reference. 

\subsubsection{High lateral slip}
\begin{figure}
\figspacebefore
\marginspace
\begin{center}
\includegraphics[width=\columnwidth]{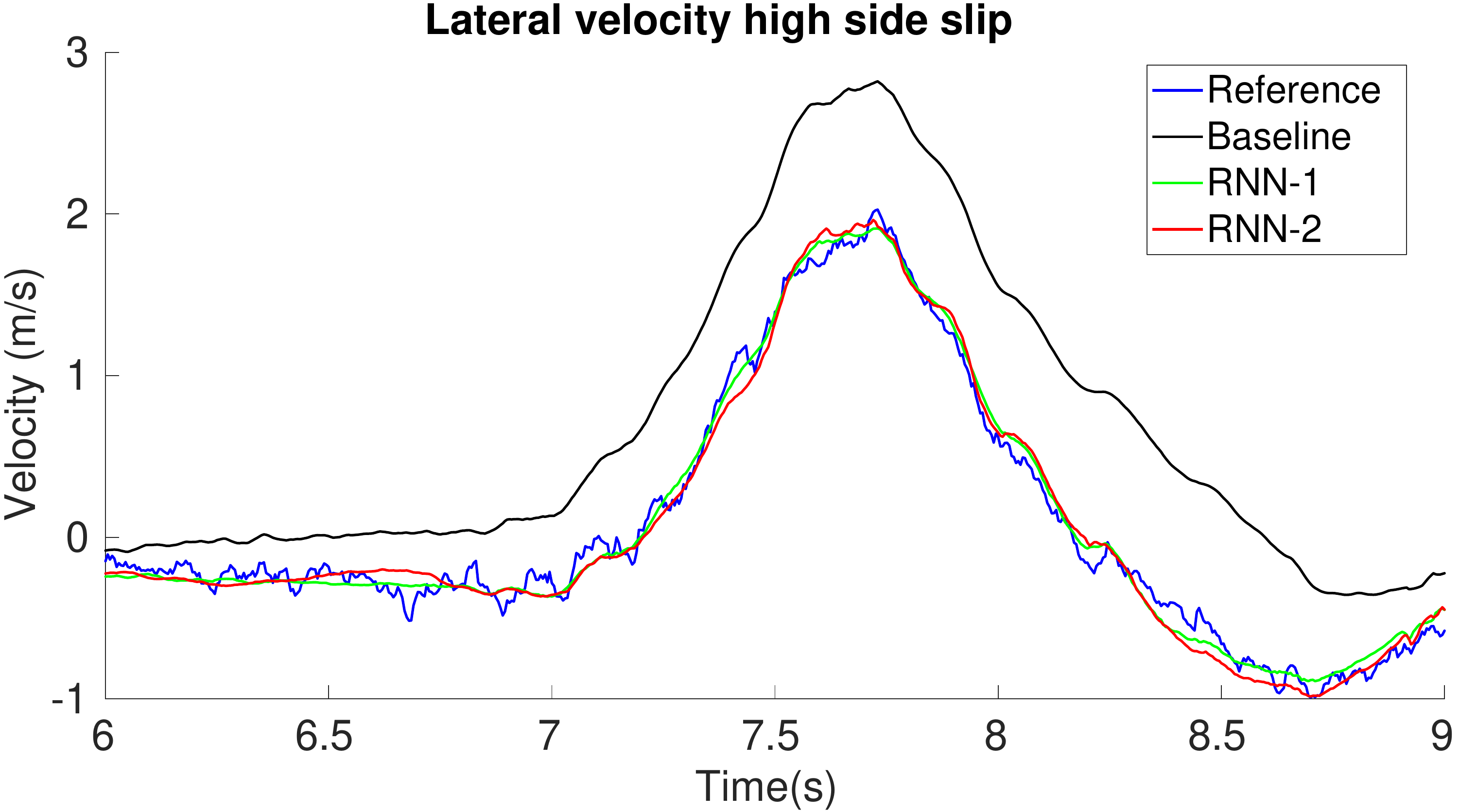}
\end{center}
    \figspacecaption
	\caption{Comparison of the lateral velocity estimate during high lateral slip condition ($\approx$ 10\degree  at the rear axle) shows the proposed recurrent neural network completely outperform the Kalman filter baseline and produce a result comparable to the reference}
	\label{fig:vy_drift}
\figspaceafter
\end{figure}

To minimise the lap-time, carrying speed through corners is essential. To maximise the cornering force, the car needs to maintain an optimal lateral slip at the tires. This is a difficult condition to control and relies on velocity feedback from velocity estimation to ensure stability. Figure \ref{fig:vy_drift} shows the lateral velocity estimate of the proposed solutions during very high lateral slip ($\approx$10\degree at the rear axle). The proposed network has learned the complex dynamics for the hardest state in a very aggressive maneuver and the estimate is very close to the reference and significantly outperforms the baseline while using the same sensor setup. 

\subsubsection{Outlier rejection}
\begin{figure}
\figspacebefore
\begin{center}
\includegraphics[width=\columnwidth]{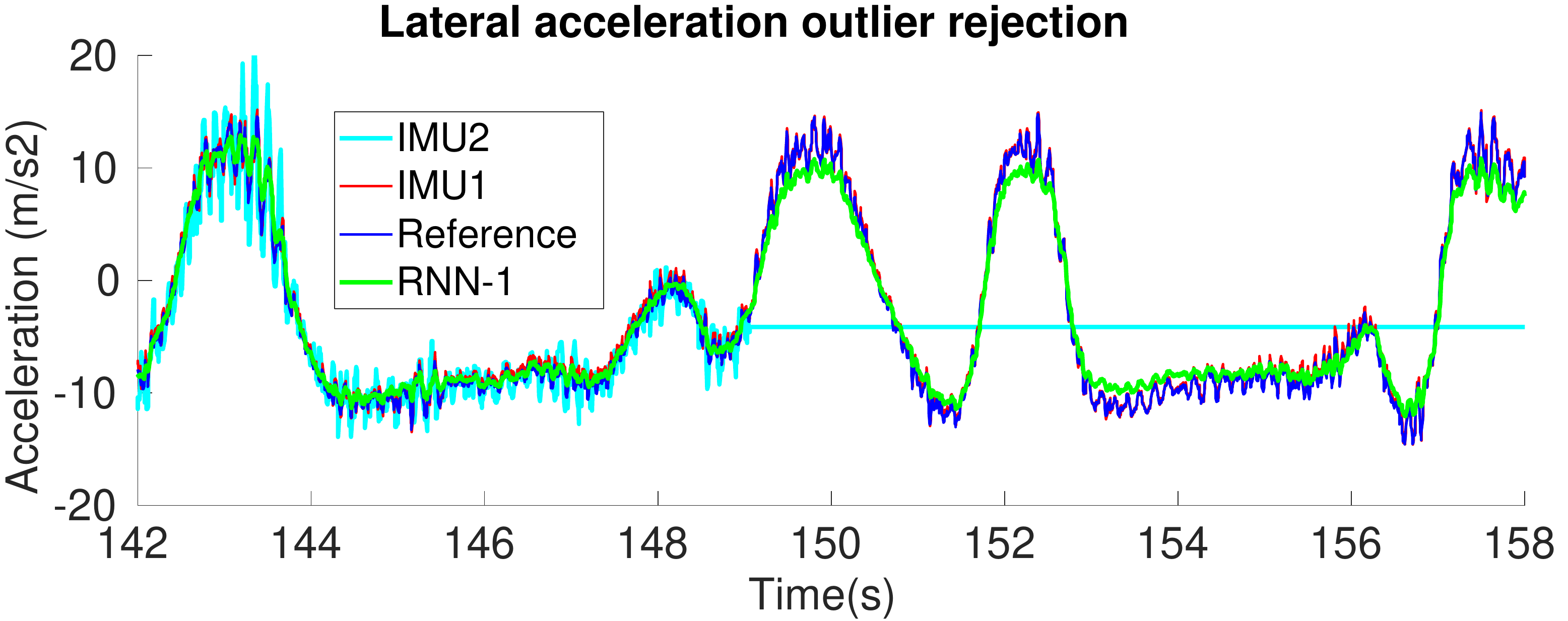}
\end{center}
    \figspacecaption
	\caption{The proposed network is able to reject bad measurements and provide a very good estimate even under very aggressive driving conditions, validated in the lateral acceleration estimate during failure of one IMU (shown in yellow)}
	\label{fig:ay_fail_outlier}
\figspaceafter
\vspace{-8pt}
\end{figure}

The importance of rejecting outlier measurements has been discussed in detail by Valls \textit{et al.} \cite{cit:fluela_icra}. In Figure \ref{fig:ay_fail_outlier}, the measurements of IMU-2 are not updated after t $\approx$ 149s. In this case, the Kalman filter has explicit outlier detection that rejects the bad measurement to provide a good estimate of the acceleration. The proposed network is capable of detecting this based on the learnt dynamics and the IMU-1 measurement resulting in the outlier measurement being implicitly rejected to match the reference. 

\section{Discussion and limitations}
\label{sec:discussion}An advantage of Kalman filters, as opposed to the RNNs, is the notion of uncertainty. Having an estimate of the uncertainty could be useful for downstream modules, but is not possible with the current architecture. Gal and Ghahramani \cite{cit:probability_dropout} suggest using dropout to obtain a Monte Carlo distribution. Mixture density functions described by Bishop \cite{cit:mixture_density} could also be used.
Another limitation would be the inability to incorporate minor, known changes. Re-training with new data is currently the only option.

\section{Conclusion}\label{sec:conclusion}We present a novel end-to-end recurrent neural network application that takes affordable and available raw sensors as input (IMU, wheel odometry, and motor currents) and estimates the output velocity. This network is extensively tested on real-world data generated while racing autonomously, evaluated at very high sideslip (10\degree\ at the rear axle), close to the limits of handling. The network outperforms the state-of-the-art Kalman filter with equivalent input data and matches the Kalman filter with expensive external velocity sensors.

\section*{ACKNOWLEDGMENT}

The authors thank the AMZ Driverless team for their sustained hard-work and passion, as well as the sponsors for their financial and technical support. We also express our gratitude to everyone at the Autonomous Systems Lab of ETH Z\"urich for their support throughout the project.


\bibliographystyle{IEEEtran}
\bibliography{root.bbl}
\end{document}